\documentclass{article}

\usepackage{microtype}
\usepackage{graphicx}
\usepackage{subfigure}
\usepackage{booktabs} 

\usepackage{hyperref}

\usepackage[accepted]{icml2024}

\usepackage{amsmath}
\usepackage{amssymb}
\usepackage{mathtools}
\usepackage{amsthm}

\usepackage[capitalize,noabbrev]{cleveref}

\theoremstyle{plain}

\theoremstyle{definition}

\theoremstyle{remark}

\usepackage[textsize=tiny]{todonotes}

\icmltitlerunning{Efficient Learning of Convolution Weights as Gaussian Mixture Model Posteriors}

\begin{document}

\twocolumn[
\icmltitle{Efficient Learning of Convolution Weights as Gaussian Mixture Model Posteriors}



\icmlsetsymbol{equal}{*}

\begin{icmlauthorlist}
\icmlauthor{Lifan Liang}{yyy}
\end{icmlauthorlist}

\icmlaffiliation{yyy}{Department of Human Genetics, University of Chicago, Illinois, United States}

\icmlcorrespondingauthor{Lifan Liang}{lifanl@uchicago.edu}

\icmlkeywords{Machine Learning, Generative model, }

\vskip 0.3in
]



\printAffiliationsAndNotice{} 

\begin{abstract}
In this paper, we showed that the feature map of a convolution layer is equivalent to the unnormalized log posterior of a special kind of Gaussian mixture for image modeling. Then we expanded the model to drive diverse features and proposed a corresponding EM algorithm to learn the model. Learning convolution weights using this approach is efficient, guaranteed to converge, and does not need supervised information. Code is available at: \url{https://github.com/LifanLiang/CALM}.
\end{abstract}

\section{Introduction}
\label{intro}
Unsupervised learning of image representations has been the focus of computer vision research for decades. In recent years, the discriminative approach has shown great promise in a series of benchmarking tasks \cite{chen2020simple,donahue2014decaf,doersch2015unsupervised,caron2020unsupervised,van2020scan}. However, the generality of such learned representation would be limited by the scope of the discriminative tasks. Meanwhile, generative representation learning on pixel level is computationally expensive \cite{lee2009convolutional}.  

In this paper, we deviated from the current trend, traced back to the generative approach to learning latent features, and proposed a patch-level generative model for s single convolution layer. We reinterpreted the convolution layer as computing posterior membership for our generative model. The corresponding training algorithm only relies on the operation of convolution without gradient computation and is guaranteed to converge. Our work provides both theoretical and practical insights for future research on unsupervised representation learning related to convolution neural network.

\section{Related Work}

To our knowledge, existing works have not yet proposed to learn convolution weights by using the corresponding feature map. Restricted Boltzmann Machine (RBM) has a similar learning approach since the latent features can be learned from the output activation \cite{hinton2002training}.

In terms of the generative model we developed, the most similar existing method is using Kmeans clustering on every patch that can be extracted from images. This method has performed favorably compared to other more sophisticated clustering methods such as RBM \cite{coates2011analysis}. This benchmark observation cannot be well explained. Our model formulation suggests that Kmeans extracts features invariant to original pixel intensities when images are normalized with variance as one.

Another important difference is that Kmeans clustering on all the image patches is computationally expensive in both time complexity and space complexity, while our training algorithm can be easily implemented in existing tools (i.e. pytorch, tensorflow) that enable GPU parallelization. In addition, the batch size will not affect model convergence.

A more recent development is Convolutional Kmeans \cite{dundar2015convolutional}, which encouraged Kmeans to avoid redundant centroids by increasing the diversity of the image patches. For our model, weight penalty combined with normalization of feature map across filters is sufficient to drive diverse convolution weights.

\section{Model Formulation}

\subsection{The generative model with a single patch}
An image with height $H$, width $W$, and $C$ channels is a patch embedded in random background. Let $(Z_1, Z_2)$ denote the top left coordinate of the patch. We assign a Multinomial distribution to $Z$ such that the patch is equally likely to be anywhere in an image. Assuming the patch is square size with length $L$, the patch area would be
$$
U = \{i,j| Z_1 \leq i \leq Z_1+L-1, Z_2 \leq j \leq Z_2+L-1 \}
$$
Pixel intensities in the background area are sampled from the standard Normal distribution, namely:
$$
X_{\{i,j,c| (i,j) \notin U\}} \sim Normal(0,1)
$$
where $X_{ijc}$ is pixel intensity located at $i$th row, $j$th column, and $c$th channel of the image. And the pixel intensities in the patch area follow a Multivariate Normal:
$$
X_{\{i,j,c| (i,j) \in U\}} \sim MVN(\mu, \mathcal{I})
$$
where U is the set of pixel locations within the patch area, $\mu$ is a vector of size $L \times L \times C$, representing the mean for every patch pixel in every channel, $\mathcal{I}$ is the identify matrix for covariance. We set all variance to be one for model simplicity. Unit variance also makes it easier to see the connection between our image modeling and convolution.

With the generative model stated above, the likelihood function regarding $U$ and $\mu$ is:
\begin{equation}
    P(X|\mu,U) \propto exp(-0.5\sum_{U}^{i,j} \mu_{i,j}^2 + \sum_U^{i,j} \mu_{i,j} X_{i,j})
\end{equation}
The log posterior of patch location, $U$ can be computed as:
\begin{equation}
\begin{aligned}
    \log P(U|X,\mu) &= \log [\dfrac{P(X|\mu,U)P(U)}{\sum_U{P(X|\mu,U)P(U)}}] \\
    &= \sum_U^{i,j} \mu_{i,j} X_{i,j} - C
\end{aligned}
\end{equation}
where $C$ is the log total sum of $\sum_U^{i,j} \mu_{i,j} X_{i,j}$ across patch locations. Please note that the term $\sum_U^{i,j} \mu_{i,j} X_{i,j}$ is identical to applying a convolution filter to location $U$ of an image. Therefore, the feature map generated from a single convolutional layer is effectively the unnormalized log posterior of patch location for this model.

\subsection{The generative model with multiple patches}
Although the single patch model works well with supervised or self-supervised learning, it will result in poor performance in unsupervised learning. This is because CNN naturally deploys multiple convolution filters. Many of them will converge to the same local optima without supervised information or drop out, resulting in many redundant features.

To address this issue, the model needs to consider multiple patches altogether. Similar to a conventional Gaussian mixture, a mixture of patches reaches higher likelihood when each patch captures different edge features of an image. Suppose the model is a mixture of $A$ patches and there are $B$ possible locations of $U$, then the number of components for multiple patch model is $A \times B$. Conditioned on patch location $U$ and the patch $V$, the two models have the same likelihood function as Equation (1). But the posterior is different:
\begin{equation} \label{eq:2}
    log P(U,V|X,\mu) = \sum_{U^V}^{i,j} \mu^{V}_{i,j} X_{i,j} - 0.5\sum_{U^V}^{i,j} {\mu^{V}_{i,j}}^2 - const
\end{equation}
where $V$ is one patch in the mixture, $U^V$ is the image area covered by the patch $V$, $\mu^V$ is the parameters of the patch $V$, and $const$ denotes the log sum of likelihood for all possible $Z$ and $V$:
$$
const = log \sum_{U,V} (\sum_{U^V}^{i,j} \mu^{V}_{i,j} X_{i,j} - 0.5\sum_{U^V}^{i,j} {\mu^{V}_{i,j}}^2)
$$
Please note that the term $0.5\sum_{U^V}^{i,j}{\mu^V_{i,j}}^2$ cannot be canceled out in the multiple patch model. This is consistent with previous observations that unsupervised learning of images requires sparsity regularization.

\begin{figure}[ht]
\vskip 0.2in
\begin{center}
\centerline{\includegraphics[width=\columnwidth]{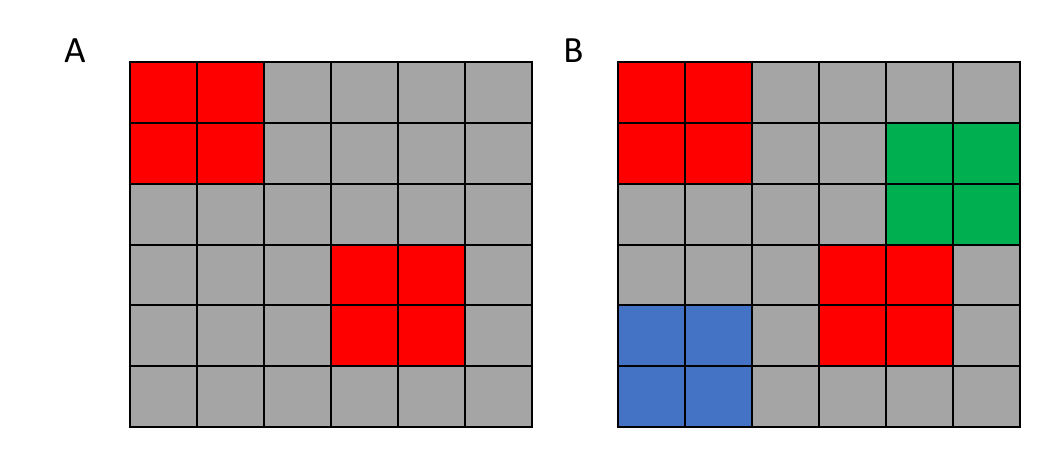}}
\caption{Graphical illustration of the single patch model and multiple patch model. 
Pixels in gray area are sampled from standard normal. 
Patches in different colors are sampled from different $MVN$}
\label{CAML1}
\end{center}
\vskip -0.2in
\end{figure}

\section{Learning Algorithm}
\subsection{Expectional maximization (EM)}
Similar to the Gaussian mixture, our model can be estimated via EM algorithm, which always increases the model likelihood and guarantees convergence to zero gradients. For the expectation step (E-step), we computed the joint posterior of $U$ and $V$, as in \cref{eq:2}. This can be implemented as conventional convolution with the term $- 0.5\sum_{U^V}^{i,j} {\mu^V_{i,j}}^2$ set to the bias.

In the maximization step (M-step), we need to estimate the mean vector for each patch. This is computed as the average across pixel intensities weighted by the joint posterior from E-step. For one parameter in a patch, the maximum likelihood estimate is:

\begin{equation} \label{eq:3}
    \hat{\mu}_{ncijk} = \dfrac{\sum_{n,i',j'} \hat{\gamma}_{nci'j'} \cdot X_{n,i'+i-1,j'+j-1,k}}{\sum_{n,i',j'} \hat{\gamma}_{nci',j'}}
\end{equation}

where $\hat{\gamma}$ is the estimated joint posterior, $n$ is the index for samples, $c$ is the index for patches, $ijk$ are indices within the height, width, and input channel ranges respectively, while $i'j'$ denotes the coordinates in $\hat{\gamma}$. When we move to estimate the element next to $\mu_{ncijk}$, we essentially apply dot product between the same $\hat{\gamma}$ and the next cover area of images. Hence the nominator in \cref{eq:3} can be computed by applying convolution using $\hat{\gamma}$ as the filter. 

However, such convolution is different from conventional convolution in two ways: (1) each image has a unique convolution filter because different images yield different $\hat{\gamma}$; (2) $\hat{\gamma}$ was applied to each color channel separately to estimate different parameters. To address the first difference with the existing deep learning framework, we took advantage of the grouped convolution method. The batch dimension was set to the input channel dimension and channels within one image make one group. As for the second difference, we used 3D convolution instead of 2D, treating the color channel as an extra dimension of the images. 

\begin{figure}[ht]
\vskip 0.2in
\begin{center}
\centerline{\includegraphics[width=\columnwidth]{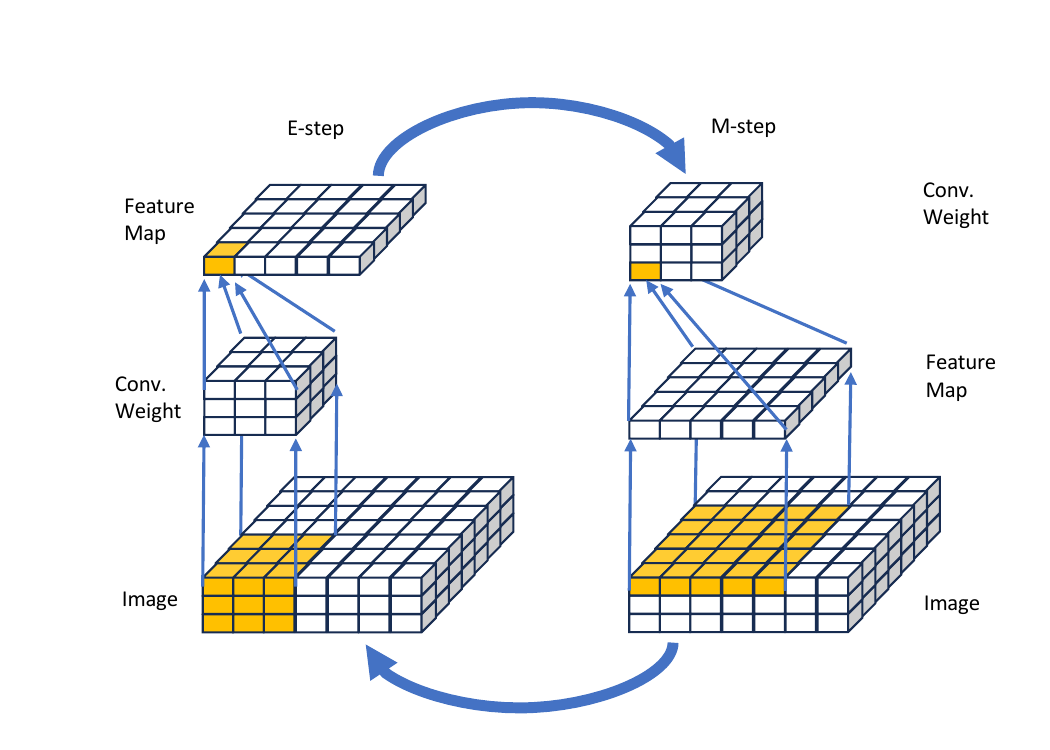}}
\caption{Simplified illustration of the EM algorithm developed in this paper.
E-step is a conventional convolution;
M-step uses the feature map as convolution filter.}
\label{EM-figure}
\end{center}
\vskip -0.2in
\end{figure}

\subsection{Convergence of the algorithm}
Treating $U$ and $V$ as the missing data and $\mu$ as the parameter, previous analysis of EM algorithm \cite{neal1998view, wu1983convergence} is applicable to our model. Essentially, we are directly optimizing the function $F$:
$$
F(U,V,\mu) = \sum_{U,V} q(U,V) \log P(X,U,V| \mu) 
$$
where $q(U,V)$ is the distribution set to $P(U,V|X, \mu)$ in E-step. The marginal likelihood, $P(X|\mu)$ will also increase as $F$ increases. 

Alternatively, our algorithm is similar to performing coordinate ascent over the function $Q$:
$$
Q(U,V,\mu) = E_q [\log P(X,U,V| \mu)] + E_q \log q(U,V)
$$

\subsection{Batch EM}
The size of $\hat{\gamma}$ does not scale well with sample size. Fortunately, we are more interested in learning convolution weights. For this purpose, running batch EM is much better. For each batch, we only need to retain the nominator and denominator of \cref{eq:3}. After one epoch, we summed up retained values across all batches and updated patch parameters according to \cref{eq:3}. \cref{alg:batchEM} describes the procedure.

\begin{algorithm}[tb]
   \caption{Batch EM}
   \label{alg:batchEM}
\begin{algorithmic}
   \STATE {\bfseries Input:} data $X$, sample size $N$, number of patches $K$, patch size $L$
   \STATE Initialize $batches = SequentialBatch(X)$.
   \STATE Initialize $\mu = Array[K, 3, L, L]$.
   \REPEAT
   \STATE Initialize $\mu^* = 0, \beta^* = 0$
   \FOR{$b$ \textbf{in} $batches$}
   \STATE $\gamma_b = Estep(X[b], \mu)$
   \STATE $\mu_b, \beta_b = BatchMstep(X[b], \gamma_b)$
   \STATE $\mu^* = \mu^* + \mu_b; \beta^* = \beta^* + \beta_b$
   \ENDFOR
   \STATE $\mu = UpdateParameters(\mu^*, \beta^*)$
   \UNTIL{$\mu$ does not change}
\end{algorithmic}
\end{algorithm}

\subsection{Image-wide pooling}
Feature maps learned from our model still contain position information. In principle, image classification should only be concerned with the activation of features, regardless of the location of activation. We summarize the feature map by integrating out the position variable from the feature map. More specifically, we performed log-sum-exp across the entire feature map for each channel.

\section{Result}
\subsection{Application to MNIST}
\begin{figure}[ht]
\vskip 0.2in
\begin{center}
\centerline{\includegraphics[width=\columnwidth]{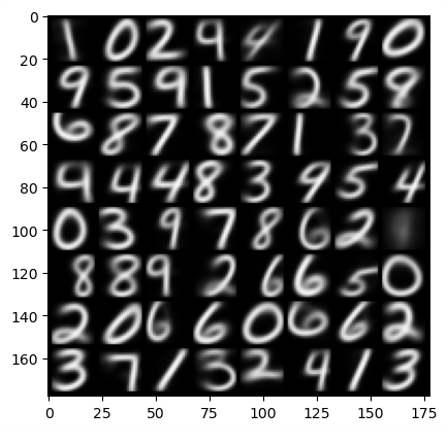}}
\caption{Visualization of the 64 features extracted from the MNIST images.}
\label{mnist-features}
\end{center}
\vskip -0.2in
\end{figure}

We applied our model to the MNIST dataset for handwritten digit recognition. Pixel intensities were divided by 255 so that they were between 0 and 1. We used 64 convolution filters with the size $20 \times 20$. Shown in \cref{mnist-features} are the 64 features learned from images. We further applied image-wide pooling on the feature map, such that every image yields a 64-dimension vector indicating feature activation. We trained a Softmax regression with this feature activation vector and image labels within the training set.

For comparison, we trained a supervised model with the same number of parameters. The supervised model also has one convolution layer with 64 convolution filters of size $20 \times 20$, followed by image-wide pooling, Relu activation, and Softmax output. As shown in \cref{accuracy}, our model performed slightly worse than the supervised counterpart. This is probably because supervised learning can identify image features more related to the task.

\begin{figure}[ht]
\vskip 0.2in
\begin{center}
\centerline{\includegraphics[width=\columnwidth]{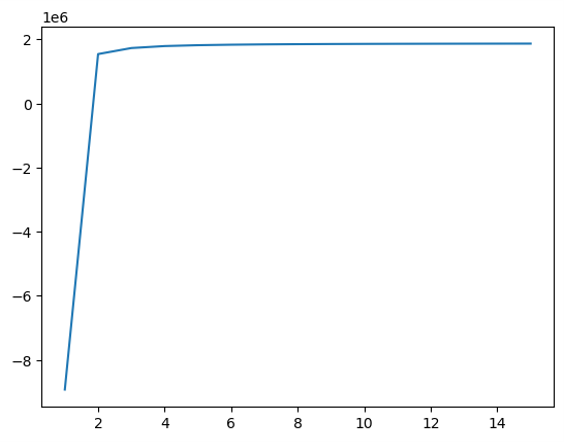}}
\caption{Unnormalized marginal likelihood along with epochs on the MNIST dataset. EM algorithm converged after the second epoch.}
\label{mnist-convergence}
\end{center}
\vskip -0.2in
\end{figure}

\begin{table}[t]
\caption{Classification accuracies for a supervised and unsupervised convolution layer. The column named "unlabeled" means that our model utilized the unlabeled data during feature extraction.}
\label{accuracy}
\vskip 0.15in
\begin{center}
\begin{small}
\begin{sc}
\begin{tabular}{lcccr}
\toprule
Data set & Supervised & Unsupervised & Unlabeled\\
\midrule
MNIST    & 96.06 & 94.82 & --\\
STL10    & 10.00 & 40.03 & 53.54\\
\bottomrule
\end{tabular}
\end{sc}
\end{small}
\end{center}
\vskip -0.1in
\end{table}

\subsection{Application to STL-10}
\begin{figure}[ht]
\vskip 0.2in
\begin{center}
\centerline{\includegraphics[width=\columnwidth]{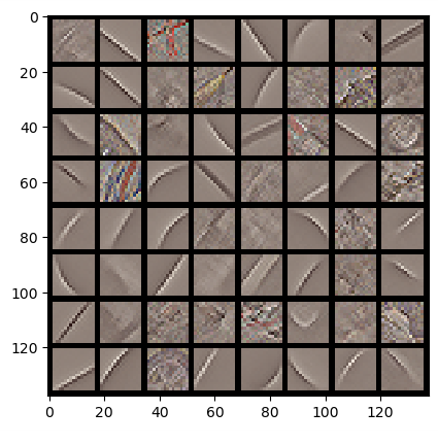}}
\caption{Visualization of the first 64 features extracted from the STL10 images (labeled training data and unlabeled data).}
\label{stl10-features}
\end{center}
\vskip -0.2in
\end{figure}

\begin{figure}[ht]
\vskip 0.2in
\begin{center}
\centerline{\includegraphics[width=\columnwidth]{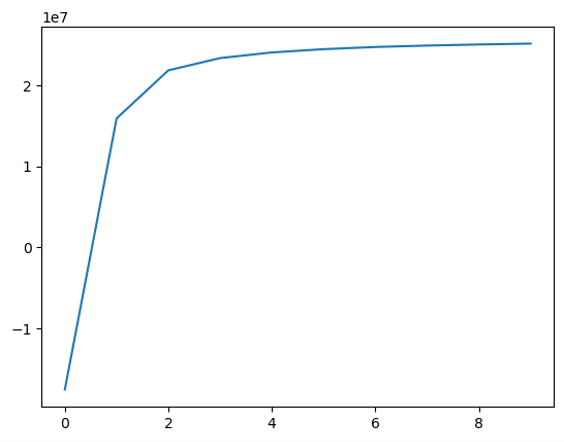}}
\caption{Unnormalized marginal likelihood along with epochs on the STL10 dataset  (labeled training data and unlabeled data).}
\label{stl10-convergence}
\end{center}
\vskip -0.2in
\end{figure}

We evaluated our model performance on STL-10 compared to a supervised counterpart. We used 200 convolution kernels of size $15 \times 15$. The rest is similar to MNIST. \cref{stl10-convergence} showed that our learning algorithm converged after 5 epochs. Most features extracted are edges (\cref{stl10-features}). 

As shown in the second row of \cref{accuracy}, the supervised model failed to learn meaningful features from the 5000 labeled images, while our model still managed to extract some features related to the labels. When labeled training data is combined with unlabeled data, our model performance has increased more than 10\%.

\section{Conclusion}
In this work, we reinterpreted a convolution layer as a special type of Gaussian mixture. The corresponding EM algorithm can be implemented as alternating the kernels for convolution, which is very simple and fast. The major limitation is that our model only allows training of a single convolution layer. A straightforward solution would be stacking these layers together to construct deep neural networks. But computational cost increased as the number of kernels increased. We leave the issue of constructing a hierarchical model to future research. 

Our training algorithm performs convolution with feature map instead of backward propagation. Although running speed may not be much faster, this approach has several advantages: (1) It enables training without any labels or self-supervision; (2) There is no need for gradient computation; (3) It strictly increases model likelihood after each epoch; (4) It does not require any parameter tuning for optimization. In terms of model performance, our experiments suggest supervised learning and unsupervised learning excel in different scenarios. When there are sufficient labels or supervised information, supervised learning is more effective because it focuses on discriminative image features. When labels are scarce compared to the complexity of images, supervised learning may struggle while unsupervised learning can still extract effective features from images. In addition, unsupervised learning can directly utilize unlabeled data to enhance performance.

\bibliography{main}

\begin{thebibliography}{11}
\providecommand{\natexlab}[1]{#1}
\providecommand{\url}[1]{\texttt{#1}}
\expandafter\ifx\csname urlstyle\endcsname\relax
  \providecommand{\doi}[1]{doi: #1}\else
  \providecommand{\doi}{doi: \begingroup \urlstyle{rm}\Url}\fi

\bibitem[Caron et~al.(2020)Caron, Misra, Mairal, Goyal, Bojanowski, and Joulin]{caron2020unsupervised}
Caron, M., Misra, I., Mairal, J., Goyal, P., Bojanowski, P., and Joulin, A.
\newblock Unsupervised learning of visual features by contrasting cluster assignments.
\newblock \emph{Advances in neural information processing systems}, 33:\penalty0 9912--9924, 2020.

\bibitem[Chen et~al.(2020)Chen, Kornblith, Norouzi, and Hinton]{chen2020simple}
Chen, T., Kornblith, S., Norouzi, M., and Hinton, G.
\newblock A simple framework for contrastive learning of visual representations.
\newblock In \emph{International conference on machine learning}, pp.\  1597--1607. PMLR, 2020.

\bibitem[Coates et~al.(2011)Coates, Ng, and Lee]{coates2011analysis}
Coates, A., Ng, A., and Lee, H.
\newblock An analysis of single-layer networks in unsupervised feature learning.
\newblock In \emph{Proceedings of the fourteenth international conference on artificial intelligence and statistics}, pp.\  215--223. JMLR Workshop and Conference Proceedings, 2011.

\bibitem[Doersch et~al.(2015)Doersch, Gupta, and Efros]{doersch2015unsupervised}
Doersch, C., Gupta, A., and Efros, A.~A.
\newblock Unsupervised visual representation learning by context prediction.
\newblock In \emph{Proceedings of the IEEE international conference on computer vision}, pp.\  1422--1430, 2015.

\bibitem[Donahue et~al.(2014)Donahue, Jia, Vinyals, Hoffman, Zhang, Tzeng, and Darrell]{donahue2014decaf}
Donahue, J., Jia, Y., Vinyals, O., Hoffman, J., Zhang, N., Tzeng, E., and Darrell, T.
\newblock Decaf: A deep convolutional activation feature for generic visual recognition.
\newblock In \emph{International conference on machine learning}, pp.\  647--655. PMLR, 2014.

\bibitem[Dundar et~al.(2015)Dundar, Jin, and Culurciello]{dundar2015convolutional}
Dundar, A., Jin, J., and Culurciello, E.
\newblock Convolutional clustering for unsupervised learning.
\newblock \emph{arXiv preprint arXiv:1511.06241}, 2015.

\bibitem[Hinton(2002)]{hinton2002training}
Hinton, G.~E.
\newblock Training products of experts by minimizing contrastive divergence.
\newblock \emph{Neural computation}, 14\penalty0 (8):\penalty0 1771--1800, 2002.

\bibitem[Lee et~al.(2009)Lee, Grosse, Ranganath, and Ng]{lee2009convolutional}
Lee, H., Grosse, R., Ranganath, R., and Ng, A.~Y.
\newblock Convolutional deep belief networks for scalable unsupervised learning of hierarchical representations.
\newblock In \emph{Proceedings of the 26th annual international conference on machine learning}, pp.\  609--616, 2009.

\bibitem[Neal \& Hinton(1998)Neal and Hinton]{neal1998view}
Neal, R.~M. and Hinton, G.~E.
\newblock A view of the em algorithm that justifies incremental, sparse, and other variants.
\newblock In \emph{Learning in graphical models}, pp.\  355--368. Springer, 1998.

\bibitem[Van~Gansbeke et~al.(2020)Van~Gansbeke, Vandenhende, Georgoulis, Proesmans, and Van~Gool]{van2020scan}
Van~Gansbeke, W., Vandenhende, S., Georgoulis, S., Proesmans, M., and Van~Gool, L.
\newblock Scan: Learning to classify images without labels.
\newblock In \emph{European conference on computer vision}, pp.\  268--285. Springer, 2020.

\bibitem[Wu(1983)]{wu1983convergence}
Wu, C.~J.
\newblock On the convergence properties of the em algorithm.
\newblock \emph{The Annals of statistics}, pp.\  95--103, 1983.

\end{thebibliography}
\bibliographystyle{icml2024}

\newpage
\appendix
\onecolumn
\end{document}